\documentclass[preprint,1p,times]{elsarticle}

\usepackage{amssymb}
\usepackage{amsmath,amsfonts}
\usepackage{algorithmic}
\usepackage{algorithm}
\usepackage{array}
\usepackage[caption=false,font=small,labelfont=sf,textfont=sf]{subfig}
\usepackage{textcomp}
\usepackage{stfloats}
\usepackage{url}
\usepackage{verbatim}
\usepackage{graphicx}
\usepackage{svg}
\usepackage{booktabs}
\usepackage{multirow}
\usepackage{xcolor}
\usepackage{threeparttable}
\usepackage{natbib}

\colorlet{softred}  {red!70!black}
\colorlet{softgreen}{green!60!black}
\colorlet{softblue} {blue!100!white}

\newcommand{\first} [1]{\textbf{\textcolor{softred}  {#1}}}
\newcommand{\second}[1]{\textbf{\textcolor{softgreen}{#1}}}
\newcommand{\third} [1]{\textbf{\textcolor{softblue} {#1}}}

\journal{Knowledge-based Systems}

\begin{document}

\begin{frontmatter}

\title{Grasp the Graph (GtG)~2.0:\\ Ensemble of Graph Neural Networks for High‑Precision Grasp Pose Detection in Clutter}

\author{Ali Rashidi~Moghadam\fnref{fn_equal}}
\author{Sayedmohammadreza~Rastegari\fnref{fn_equal}}
\author{Mehdi~Tale~Masouleh\corref{cor1}}
\author{Ahmad~Kalhor}

\address{Human and Robot Interaction Laboratory, School of Electrical and Computer Engineering, University of Tehran, Tehran, Iran}

\fntext[fn_equal]{Equal contribution.}

\cortext[cor1]{Corresponding author: m.t.masouleh@ut.ac.ir}

\begin{abstract}
This paper presents \emph{Grasp the Graph~2.0 (GtG~2.0)}, an effective two-stage framework for 7-DoF grasp pose detection that leverages an ensemble of Graph Neural Networks (GNNs) for geometric reasoning on point cloud data. Building on the success of GtG~1.0 which demonstrated the feasibility of GNN-based 4-DoF grasp detection in simulation, GtG~2.0 integrates a conventional Grasp Pose Generator to generate diverse 7-DoF grasp candidates. These candidates are then scored using an ensemble of lightweight GNN-based regressors.
Each model processes a local region around the candidate by incorporating both points between the gripper jaws (\emph{inside points}) and points from the surrounding region (\emph{outside points}). This enriched representation enhances robustness against occlusion and partial views, improving grasp score prediction.
GtG~2.0 achieves state-of-the-art performance among all hypothesis-and-test and GNN-based methods on the GraspNet-1Billion benchmark and ranks third overall across all evaluated frameworks. It demonstrates up to a 35\% improvement in Average Precision compared to similar approaches. Furthermore, real-world experiments on a 4-DoF grasping setup equipped with a Kinect-v1 camera confirm its flexibility and reliability, with a 91\% grasp success rate and 100\% scene completion in cluttered scenarios. The source code is available at \url{https://github.com/Ali-Rashidi/GtG2}.
\end{abstract}


\end{frontmatter}

\section*{Highlights}
\begin{itemize}
\item GtG~2.0 uses localized inside/outside graph representations to evaluate 7-DoF grasps.
\item An ensemble of GNNs boosts grasp AP by up to 35\% on GraspNet-1Billion.
\item Real robot experiments achieve 91\% grasp success and 100\% task completion.
\item Requires fewer parameters than similar methods.
\item Generalizes across different configurations, tested on 4-DoF after 7-DoF training.
\end{itemize}

\section{Introduction}\label{sec:introduction}

Despite decades of progress in robotic manipulation, reliably grasping objects in cluttered, unstructured environments remains an open challenge. While tasks like picking a coffee mug from a messy desk are intuitive for humans, they require robots to perform complex spatial reasoning under significant uncertainty. These challenges arise from complex object geometries, occlusions, sensor noise, and partial point cloud data.
Traditional grasp detection methods often rely on analytical metrics that evaluate the geometric relationship between the gripper’s contact points and the object~\cite{siciliano2016robotics}. These approaches typically assume access to an accurate and complete object model, which is rarely available when dealing with novel or partially observed objects in real-world scenarios.
To overcome this limitation, data-driven techniques have emerged, enabling robots to predict feasible grasp poses directly from raw sensor data, such as point clouds, without requiring explicit object models. 

Given the inherently geometric nature of grasping, {Graph Neural Networks} (GNNs)~\cite{khoshraftar2024survey, velivckovic2023everything} as a tool for geometric learning, provide a strong inductive bias for reasoning about spatial relationships within point cloud data.
Building on this insight, a line of research known as \emph{Grasp the Graph (GtG)} was introduced, which leverages GNNs to evaluate\textit{ grasp} poses using \textit{graph}-based representations of point clouds.
The initial study, GtG~1.0~\cite{10412387}, demonstrated the use of GNNs for grasp detection in a reinforcement learning setting within a simulated environment. However, the framework faced several critical limitations that restricted its applicability to real-world scenarios. First, it relied on the assumption of complete and noise-free point clouds, which are rarely available in practice due to sensor noise and occlusions. Second, it used uniform random sampling to generate 4-DoF grasp candidates—a strategy that is inefficient in cluttered scenes and unsuitable for higher-degree-of-freedom grasping tasks. Third, it processed the entire scene graph to evaluate a single grasp pose, introducing unnecessary computational overhead and making it difficult for the model to learn precise associations between a grasp candidate and its relevant local geometry.

This paper introduces \emph{Grasp the Graph 2.0 (GtG~2.0)} to address the limitations of previous work in a hypothesis-and-test setting. This approach preserves the strengths of GNN-based geometric reasoning while incorporating several key improvements for real-world applicability. First, it employs a well-studied and reliable Grasp Pose Generator (GPG)~\cite{gualtieri_high_2016} to generate 7-DoF grasp candidates directly from raw point cloud data. For each candidate, a local region is extracted from the point cloud, consisting of two parts: (1) \emph{inside points}, which lie between the gripper fingers and provide information about the object surfaces in the grasp region; and (2) \emph{outside points}, which are sampled from the surrounding area and offer broader geometric context, including cues about nearby surfaces and potential collisions. For each grasp candidate, a graph representation of the inside–outside region is constructed, and an ensemble of GNNs is employed to predict grasp score. The model was trained using labeled 7-DoF grasp poses generated on the GraspNet-1Billion~\cite{fang2020graspnet} benchmark training scenes.

In order to evaluate the performance of the proposed method, extensive experiments were conducted in both simulation and real-world settings. On the GraspNet-1Billion benchmark, GtG~2.0 establishes a new state of the art among hypothesis-and-test and GNN-based methods, improving Average Precision (AP) by up to 35\% over previous approaches, while requiring substantially fewer parameters. Furthermore, among all published methods to date, it ranks third overall on this benchmark.
For real-world evaluation, the method is tested in a 4-DoF setting and demonstrates a 91\% grasp success rate and a 100\% task completion rate across various cluttered scenes. In addition to its strong performance, the method also exhibits flexibility across different degrees of freedom at inference time, achieving reliable results even when trained with different DoF configurations.
In summary, the contributions of this paper are:
\begin{itemize}
    \item Introducing \emph{Grasp the Graph 2.0 (GtG~2.0)}, a novel grasp evaluation framework that combines a hypothesis-and-test strategy with graph-based learning.

    \item Proposing a localized graph construction method based on \emph{inside} and \emph{outside} point regions, allowing the model to jointly reason about contact geometry and broader spatial context.

    \item Employing an ensemble of Graph Neural Networks to improve robustness and generalization, enabling reliable 7-DoF grasp evaluation in cluttered and partially observed environments.

    \item Achieving state-of-the-art performance among hypothesis-and-test and GNN-based methods on the GraspNet-1Billion benchmark, with up to a 35\% improvement in Average Precision while using significantly fewer parameters; ranks third overall among all published methods.

    \item Demonstrates strong real-world performance in 4-DoF grasping scenarios, attaining a 91\% grasp success rate and 100\% task completion rate, and generalizes effectively to different DoF configurations at inference time.
\end{itemize}

The remainder of the paper is organized as follows. In Section~\ref{sec:related_work}, a comprehensive review of methods in robotic grasping is provided, with both conventional methods and recent advances, including ones employing GNNs. In Section~\ref{sec:methodology}, the proposed approach is described in detail, and the design and implementation of the GtG~2.0 framework are presented—from the grasp candidate generation and graph representation construction to the ensemble GNN-based scoring mechanism and training strategies. In Section~\ref{sec:experiments}, an extensive evaluation of the method is presented, including benchmark comparisons, ablation studies, and real-world experiments that demonstrate the robustness and precision of the framework. Finally, in Section~\ref{sec:conclusion}, the key findings are summarized, the implications of the research are discussed, and promising directions for future work in grasp detection and manipulation in cluttered environments are outlined.

\section{Related Work}
\label{sec:related_work}

Advances in data-driven grasp pose detection have led to a variety of methods, which can be broadly categorized into \textit{hypothesize-and-test} and \textit{end-to-end} approaches~\cite{platt_grasp_2023}. These paradigms differ in their approach to grasp candidate generation and evaluation, with each offering unique advantages and challenges. The following sections summarize these approaches, followed by a discussion of recent developments using GNNs in grasp detection.

\subsection{Hypothesize-and-Test Grasp Pose Detection}
In hypothesize-and-test methods, the grasp detection problem is divided into two stages: candidate generation and candidate evaluation. A candidate generator produces potential grasp poses using different heuristics, which are then assessed by a separate model which acts as an evaluator. This modular design offers flexibility, as the generator and evaluator can be optimized independently for specific tasks or sensor configurations.
For example, GQ-CNN~\cite{dx2} is a method trained on a large synthetic dataset where each depth image is paired with a parallel-jaw grasp specification and an analytic robustness metric derived from physics-based models. At inference time, the model quickly predicts the success probability of each candidate, facilitating efficient ranking and selection. Similarly, GPD~\cite{ten2017grasp} employs GPG to sample candidates, followed by a CNN-based classifier. While these methods have shown promise, they often struggle with the computational cost of processing large numbers of candidates and the challenge of generalizing to unseen objects and environments.
Improvements in handling raw point cloud data have also been explored. PointNet~\cite{charles_pointnet_2017} preserves the permutation invariance of point clouds and processes them more effectively than voxel or grid-based methods. Building on this, PointNetGPD~\cite{liang_pointnetgpd_2019} and 3DSGrasp~\cite{mohammadi20233dsgrasp} further refine grasp detection; the latter, for instance, enhances accuracy by first completing missing parts of the point cloud. UPG~\cite{upg} also proposed a framework that first uses U-disparity map analysis to classify scenes and then employs a PointNet++ based network to segment the topmost object in cluttered piles for 6-DOF pose estimation. These methods demonstrate the potential of directly processing point clouds, but they often require large datasets and computational resources, limiting their applicability.

\subsection{End-to-End Grasp Pose Detection}
End-to-end methods integrate candidate generation and evaluation within a single unified network. These approaches exploit local or global scene features to directly propose high-quality grasp candidates and assess them without the need for a separate classification stage. This integration often leads to more efficient and cohesive systems, but it can also increase the complexity of training and inference, and reduce flexibility.
For instance, RGB Matters~\cite{gou_rgb_2021} predicts gripper views and analytically computes grasp parameters such as width and depth. REGNet~\cite{zhao_regnet_2021} employs a three-stage network comprising a Score Network (SN) for grasp confidence regression, a Grasp Region Network (GRN) for proposal generation, and a Refine Network (RN) for further enhancement of proposal accuracy. These methods demonstrate the potential of end-to-end learning but often require large amounts of labeled data and computational resources.
Other approaches, such as GSNet~\cite{wang_graspness_2024} and HGGD~\cite{chen_efficient_2023}, utilize geometric cues and grasp heatmaps to guide detection. LF-GraspNet~\cite{liu2024structured} integrates a conditional VAE conditioned on structured local TSDF features to jointly learn scene geometry encoding and grasp configuration generation. SCNet~\cite{yu2023robotic} is an end-to-end category-level object pose estimation network that directly learns to deform and refine shape priors for one-shot 6-DOF pose prediction. Furthermore, \cite{hosseini2024multi} presents a grasp detection method based on object decomposition, which leverages multi-modal input (RGB and RGB-D images) and focuses on primitive shape abstraction and utilizing predefined grasp poses. More recent methods like FlexLoG~\cite{XIE2026112088} and RNGNet~\cite{chen2024regionaware} propose unified representations and integrated guidance mechanisms that further boost grasp detection accuracy. While these methods have achieved impressive results, they often rely on complex architectures that may not be suitable for resource-constrained systems and cannot be easily modified for new situations.

\subsection{Graph Neural Networks in Grasp Pose Detection}
GNNs have emerged as an effective tool for grasp detection by modeling the spatial and relational structure of point cloud data. In these methods, point clouds are represented as graphs in which nodes correspond to points or features and edges capture spatial relationships. This representation allows GNNs to efficiently reason about the geometric relationships between the gripper and objects, making them particularly well-suited for grasp detection in cluttered and dynamic environments.
One approach~\cite{9811601} formulates cluttered scenes as graphs, where nodes represent object geometries and edges encode their spatial relationships, allowing a GNN to evaluate the graph and propose feasible 6-Dof grasps for target objects. Similarly, GraNet~\cite{GraNet} constructs multi-level graphs at the scene, object, and grasp point levels, and employs a structure-aware attention mechanism to refine local relationships and improve grasp predictions. These methods demonstrate the potential of GNNs for grasp detection but often require complex architectures and large amounts of training data.
Other studies have integrated GNNs with reinforcement learning to handle dynamic or deformable objects~\cite{living_object}, dividing the task into pre-grasp and in-hand stages. The GtG~1.0 framework also leverages GNNs to process point clouds efficiently by framing grasping as a one-step reinforcement learning problem without requiring large, complex networks. Additionally, a recently introduced dual-branch GNN-based method~\cite{ZHUANG2025102879} uses one branch to learn global geometric features and another to focus on high-value grasping locations, further enhancing grasp detection performance. These approaches highlight the versatility of GNNs for reliable grasp pose detection.

\section{The GtG~2.0 Framework: Design and Implementation}
\label{sec:methodology}

\begin{figure}[t]
    \centering
    \includegraphics[width=\textwidth]{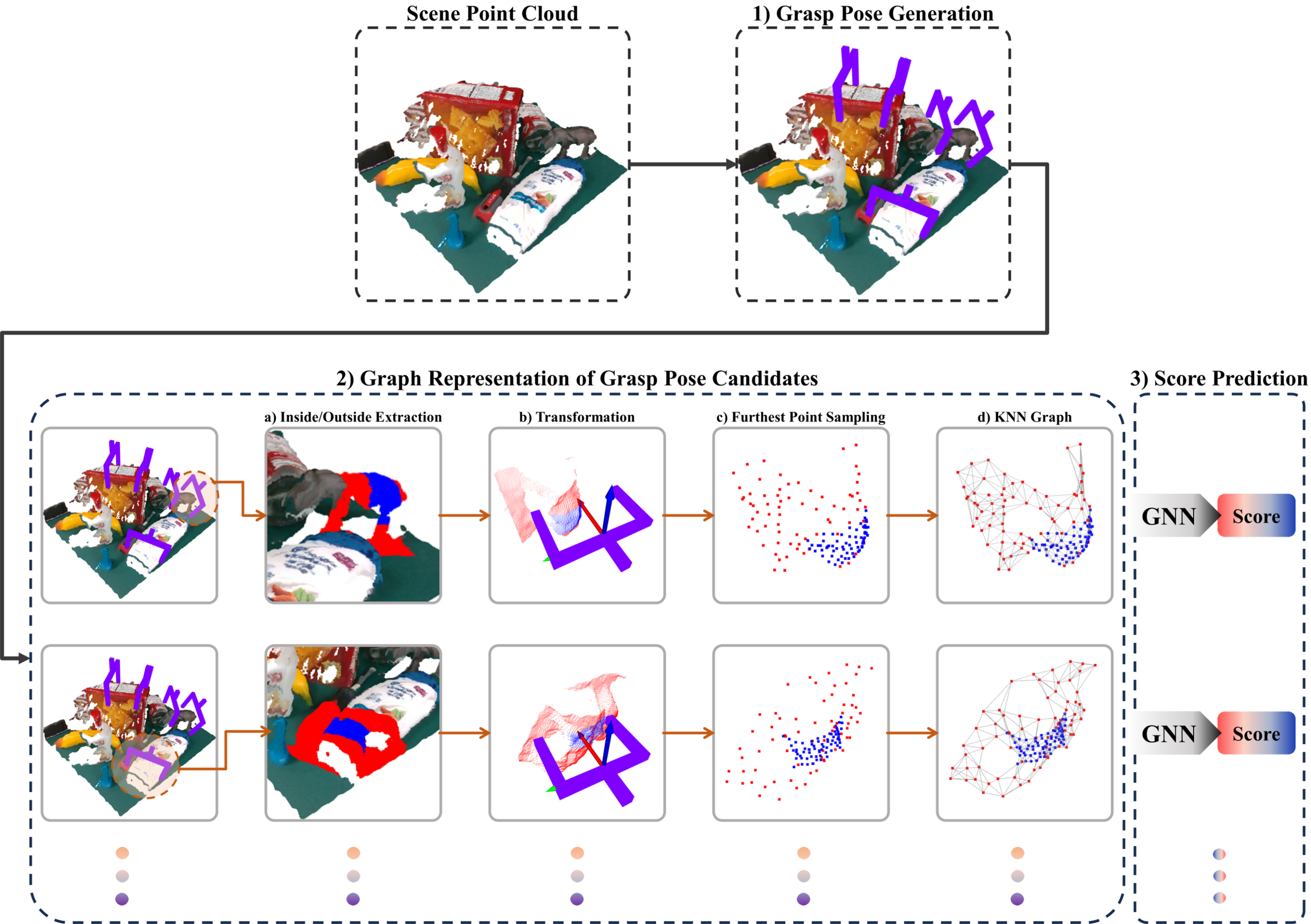}
    \caption{Overview of the GtG~2.0 pipeline. 
    \textbf{(1)} Grasp candidates are generated using the GPG algorithm. 
    \textbf{(2)} Each candidate’s local region is segmented into \textcolor{softblue}{\emph{inside points}} and \textcolor{softred}{\emph{outside points}}, which are sampled and converted into a graph using $k$-NN. 
    \textbf{(3)} The resulting graph is passed through an ensemble of GNNs. Each model predicts a grasp score, and the final score is the average of all ensemble outputs.}
    \label{fig:method}
\end{figure}

The proposed GtG~2.0 framework follows a modular, two-stage pipeline designed for 7-DoF grasp pose detection in cluttered environments. As illustrated in Figure~\ref{fig:method}, the process is composed of three main components, each playing a crucial role in extracting, encoding, and evaluating potential grasps.
First, a set of diverse grasp candidates is generated using GPG (Figure~\ref{fig:method}.1). For each candidate, a local region of the point cloud is segmented and decomposed into two subsets: \emph{inside points}, which lie between the gripper fingers, and \emph{outside points}, which provide additional information about the surroung area of the grasp. These points are then sampled and connected using $k$-nearest neighbors ($k$-NN) to form a graph (Figure~\ref{fig:method}.2). Finally, each graph is passed through an ensemble of lightweight GNNs, and their predicted grasp scores are averaged to yield a final score estimate for each pose (Figure~\ref{fig:method}.3). The entire pipeline is also detailed in Algorithm~\ref{alg:prediction_pipeline}.
The remainder of this section details each stage of the GtG~2.0 pipeline. Section~\ref{sub:m1} introduces the grasp candidate generation process. Section~\ref{subsec:m2} describes the graph representation and sampling procedure. The GNN architecture and scoring mechanism are explained in Section~\ref{sub:m3}, followed by dataset generation and training strategies in Sections~\ref{sub:m4} and~\ref{sub:m5}, respectively.

\begin{algorithm}[t]
\caption{Prediction Pipeline of GtG~2.0}
\label{alg:prediction_pipeline}
\begin{algorithmic}[1]
\STATE \textbf{Input:}  point cloud, Ensemble of trained models $\{M_1, M_2, \dots, M_5\}$
\STATE \textbf{Output:} Final grasp quality scores for each candidate grasp

\medskip
\STATE \textbf{// Step 1: Candidate Generation}
\STATE $candidates1 \leftarrow$ GPG(PointCloud)
\STATE $candidates2 \leftarrow$GPG(DepthInpainting(PointCloud))
\STATE $candidates \leftarrow  $candidates1 + candidates2
\STATE $candidates \leftarrow$ NonMaxSuppression($candidates$, $\Delta T$, $\Delta \alpha$)

\medskip
\STATE \textbf{// Step 2: Graph Representation Construction for Each Candidate}
\FOR{each candidate in $candidates$}
    \STATE Extract \textit{inside} and \textit{outside} regions from the candidate
    \STATE $inside \leftarrow$ FPS($candidate.insideRegion$, maxPoints = 70)
    \STATE $outside \leftarrow$ FPS($candidate.outsideRegion$, maxPoints = 70)
    \STATE $G \leftarrow$ ConstructKNNGraph($inside \cup outside$, $k=5$)
    \STATE Store $G$ in CandidateGraphs
\ENDFOR

\medskip
\STATE \textbf{// Step 3: Ensemble Prediction for Grasp Quality}
\FOR{each candidate graph $G$ in CandidateGraphs}
    \STATE Initialize list: \texttt{scores} $\leftarrow []$
    \FOR{each model $M$ in $\{M_1, M_2, \dots, M_5\}$}
         \STATE $s \leftarrow M.\text{Predict}(G)$ \COMMENT{// Compute grasp score for $G$}
         \STATE Append $s$ to \texttt{scores}
    \ENDFOR
    \STATE $G.final\_score \leftarrow \text{Average}(\texttt{scores})$
\ENDFOR

\medskip
\STATE \textbf{Return} CandidateGraphs with their final grasp quality scores
\end{algorithmic}
\end{algorithm}

\subsection{Grasp Pose Generation via GPG}
\label{sub:m1}

As the first stage of the GtG~2.0 framework (Figure~\ref{fig:method}.1), grasp candidates are generated using GPG, a geometry-based algorithm designed to produce 7-DoF grasp poses directly from point cloud data. GPG identifies viable grasp configurations by analyzing local surface geometry and estimating contact feasibility.
The process begins by uniformly sampling points from the input point cloud. For each sampled point, a local Darboux frame is constructed using the surface normal and principal curvature directions. A virtual parallel-jaw gripper is then aligned to this frame, with its approach vector following the surface normal. By rotating the gripper around this axis, multiple grasp orientations are created. Each configuration is simulated by moving the gripper forward along the surface normal until just before a collision occurs between the gripper body and the point cloud. Candidates that do not contain any points between the gripper fingers are then discarded.
To integrate GPG into the proposed framework, its Python implementation, \texttt{pygpg}~\cite{pygpg}, was adopted. To improve its performance under real-world conditions, several modifications were applied:

\begin{itemize}
    \item \textbf{Outlier Retention:} By default, GPG removes statistical outliers. However, this can result in grasp candidates being placed too close to the object surface, increasing the risk of collision. To avoid this, outlier removal was disabled, encouraging more conservative and stable grasp proposals.

    \item \textbf{Depth Inpainting:} Incomplete sensor coverage and occlusions often create empty regions in the point cloud. To prevent GPG from generating grasps in these unreliable areas, a Navier-Stokes-based depth inpainting algorithm \cite{Bertalmo2001NavierstokesFD} is applied. While this may introduce small artifacts, it leads to more conservative grasp generation.

    \item \textbf{Dual-Cloud Sampling with NMS:} To leverage both the original and inpainted point clouds, grasp candidates are generated on each independently. The resulting sets are merged, and Non-Maximum Suppression (NMS) is applied using thresholds of $\Delta{\text{pos}} = 5\,\text{mm}$ and $\Delta{\text{angle}} = 1^\circ$ to eliminate redundant candidates while preserving diversity.
\end{itemize}

The output of this stage is a diverse and physically plausible set of 7-DoF grasp candidates.

\subsection{Graph Representation of Grasp Pose Candidates}
\label{subsec:m2}

After generating grasp candidates, each grasp pose must be transformed into a graph representation suitable for GNN-based scoring. This process is illustrated in Figure~\ref{fig:method}.2.
For every candidate, a local region around the gripper is extracted (Figure~\ref{fig:method}.2.a), capturing two distinct sets of points: a) \emph{inside points}: those located between the gripper fingers, b) \emph{outside points}: those in the surrounding area. Unlike earlier approaches, such as GPD and PointNetGPD, which rely only on inside points, the inclusion of outside points provides additional information for reasoning about collisions and partial visibility.
All points are transformed into the local coordinate frame of the gripper to unify pose variations (Figure~\ref{fig:method}.2.b). To reduce redundancy while preserving geometric detail, Furthest Point Sampling (FPS)~\cite{fps} is applied independently to the inside and outside sets, keeping up to 70 points from each Figure~(\ref{fig:method}.2.c).
Finally, a $k$-NN graph ($k=5$) is constructed over the combined set of points (Figure~\ref{fig:method}.2.d) to enable effective feature exchange and spatial reasoning across both regions. Each node holds a 5D feature vector comprising its 3D position $(x, y, z)$ and a one-hot encoding indicating region type: $[1, 0]$ for \emph{inside} and $[0, 1]$ for \emph{outside}. Edges are created purely based on Euclidean distance, independent of the point labels.
Formally, each grasp candidate is represented as a single graph $G$, defined as a tuple:
\begin{equation}
    G = (\mathcal{V}, \mathcal{E}, \mathbf{X}),
\end{equation}
where the components are defined as follows:
\begin{itemize}
    \item $\mathcal{V} = \mathcal{V}_I \cup \mathcal{V}_O$ is the set of nodes, representing sampled points. Here, $\mathcal{V}_I$ and $\mathcal{V}_O$ are the sets of up to 70 points sampled via FPS from the \emph{inside points} and \emph{outside points}, respectively. The total number of nodes is $|\mathcal{V}| \le 140$.
    \item $\mathcal{E} = \{(v_a, v_b) \mid v_b \in \text{$k$-NN}(v_a, k=5)\}$ 
    is the set of edges. An edge exists between nodes $v_a$ and $v_b$ if $v_b$ is one of the $k=5$ nearest neighbors of $v_a$ based on the Euclidean distance between their 3D spatial coordinates.
    \item $\mathbf{X} \in \mathbb{R}^{|\mathcal{V}| \times 5}$ is the node feature matrix. Each row corresponds to a node $v_j \in \mathcal{V}$ and is represented by a 5-dimensional vector. This vector consists of its 3D spatial coordinates $(x_j, y_j, z_j)$ concatenated with a 2D one-hot encoding for its type: $[1, 0]$ for an \emph{inside point} and $[0, 1]$ for an \emph{outside point}.
\end{itemize}

\subsection{GNN-Based Grasp Pose Score Prediction}
\label{sub:m3}

Once a grasp candidate has been converted into a graph structure, the final stage of the GtG~2.0 framework involves predicting its score using a GNN-based architecture. As illustrated in Figure~\ref{fig:gnn_arch}, each input graph is processed by an ensemble of lightweight GNN-based models, which output individual grasp scores. These scores are then averaged to produce the final prediction, ensuring robustness across diverse scene configurations and input noise.

The architecture of the GNN-based model, shown in Figure~\ref{fig:gnn_arch}, is composed of four main components. First, node features, comprising 3D coordinates and a one-hot encoding indicating inside/outside labels, are processed by a Position \& Label Encoder. Next, the encoded features are passed through a stack of three SAGEConv layers to enable spatial message passing across the graph. The resulting node embeddings are then refined using an Element-wise Transformation block. These per-node features are aggregated via Global Max Pooling to obtain a fixed-size vector, which is finally passed through a Predictor MLP to generate the score. The ensemble, which will be discussed in Section~\ref{sub:m5} with more detail, consists of five such networks, each trained independently with different random seeds. The formal detail of each of the models is as follows:

\begin{figure}[t]
    \centering
    \includegraphics[width=\textwidth]{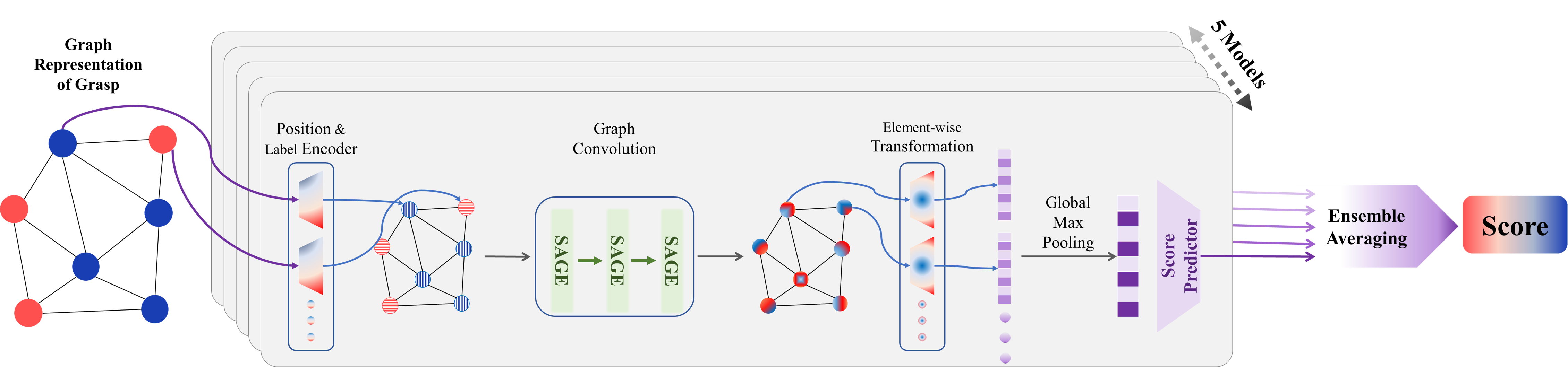}
    \caption{Architecture of the GNN-based grasp scoring network. Each input graph, composed of \emph{\textcolor{softblue}{inside points}} and \emph{\textcolor{softred}{outside points}}, is passed through a position and label encoder, a stack of SAGEConv layers, and an element-wise transformation block. Node-level features are aggregated via global max pooling and fed into a final predictor MLP to generate a score. An ensemble of five such GNNs processes the same graph independently, and their outputs are averaged to produce the final grasp score.}
    \label{fig:gnn_arch}
\end{figure}

\noindent
Each feature of the candidate graph's nodes is first encoded into a higher-dimensional representation using an encoding Multi-Layer Perceptron (MLP). Let \(\mathbf{x}_i \in \mathbb{R}^{d_{\text{in}}}\) represent the initial feature vector of node \(i\). The encoding MLP transforms \(\mathbf{x}_i\) through a series of linear mappings, batch normalizations, and ReLU activations, as follows:
\begin{equation}
\label{eq:encoding-mlp}
\begin{aligned}
\mathbf{z}^{(1)}_i &= \operatorname{ReLU}\!\Bigl(\operatorname{BN}_1(\mathbf{W}_1 \mathbf{x}_i)\Bigr),\\
\mathbf{z}^{(2)}_i &= \operatorname{ReLU}\!\Bigl(\operatorname{BN}_2(\mathbf{W}_2 \mathbf{z}^{(1)}_i)\Bigr),\\
\mathbf{z}_i &= \operatorname{BN}_3(\mathbf{W}_3 \mathbf{z}^{(2)}_i).
\end{aligned}
\end{equation}

where \(\mathbf{W}_1 \in \mathbb{R}^{d_h \times d_{\text{in}}}\), \(\mathbf{W}_2 \in \mathbb{R}^{2d_h \times d_h}\), \(\mathbf{W}_3 \in \mathbb{R}^{d_h \times 2d_h}\) are learnable weight matrices, \(d_h\) is the hidden dimension, and \(\operatorname{BN}_k\) denotes the \(k\)th batch normalization layer.

Subsequently, the graph undergoes three layers of graph convolution using the SAGEConv operator~\cite{sage}. Formally, the SAGEConv operation is defined as:
\begin{equation}
\mathbf{h}_i' = \mathbf{W}_1^{\text{sage}}\, \mathbf{h}_i + \mathbf{W}_2^{\text{sage}} \cdot \mathrm{Max}\Bigl(\{\mathbf{h}_j : j \in \mathcal{N}(i)\}\Bigr),
\end{equation}

where \(\mathbf{h}_i\) and \(\mathbf{h}_i'\) denote the input and updated feature vectors of node \(i\); \(\mathcal{N}(i)\) represents the set of neighbors of node \(i\); \(\mathbf{W}_1^{\text{sage}}\) and \(\mathbf{W}_2^{\text{sage}}\) are learnable weight matrices; and \(\mathrm{Max}(\cdot)\) computes the maximum of the features of the neighboring nodes. Following graph convolution, each final node embedding vector \(h_i\) is refined individually through an element-wise transformation:
\begin{equation}
h'_i = a(h_i) \cdot h_i + b(h_i),
\end{equation}
where \(a(h_i)\) and \(b(h_i)\) are learned vectors representing the slope and bias applied element-wise to \(h_i\). After refinement, a global max pooling operation aggregates all node embeddings into a single graph-level descriptor. This operation is expressed as:
\begin{equation}
\mathbf{h}_{\text{global}} = \max_{i \in \mathcal{V}} \{h'_i\},
\end{equation}
where \(\mathcal{V}\) denotes the set of all nodes in the graph and the maximum is computed element-wise. Finally, the global descriptor is processed by a score predictor MLP to estimate the final grasp quality score. Let \(\mathbf{h}_{\text{global}} \in \mathbb{R}^{d_h}\) be the pooled feature vector; the predictor MLP computes:
\begin{equation}
\begin{aligned}
\mathbf{s}^{(1)} &= \operatorname{ReLU}\Bigl(\operatorname{BN}_1'\bigl(\mathbf{W}_1' \mathbf{h}_{\text{global}}\bigr)\Bigr),\\[1mm]
\mathbf{s}^{(2)} &= \operatorname{ReLU}\Bigl(\operatorname{BN}_2'\bigl(\mathbf{W}_2' \mathbf{s}^{(1)}\bigr)\Bigr),\\[1mm]
s &= \mathbf{W}_3' \mathbf{s}^{(2)},
\end{aligned}
\end{equation}
where \(\mathbf{W}_1' \in \mathbb{R}^{256 \times d_h}\), \(\mathbf{W}_2' \in \mathbb{R}^{128 \times 256}\), \(\mathbf{W}_3' \in \mathbb{R}^{1 \times 128}\) are learnable weights, and \(s \in \mathbb{R}\) is the predicted grasp quality score.%

\begin{figure}[t]
\centering
\includegraphics[width=0.9\linewidth]{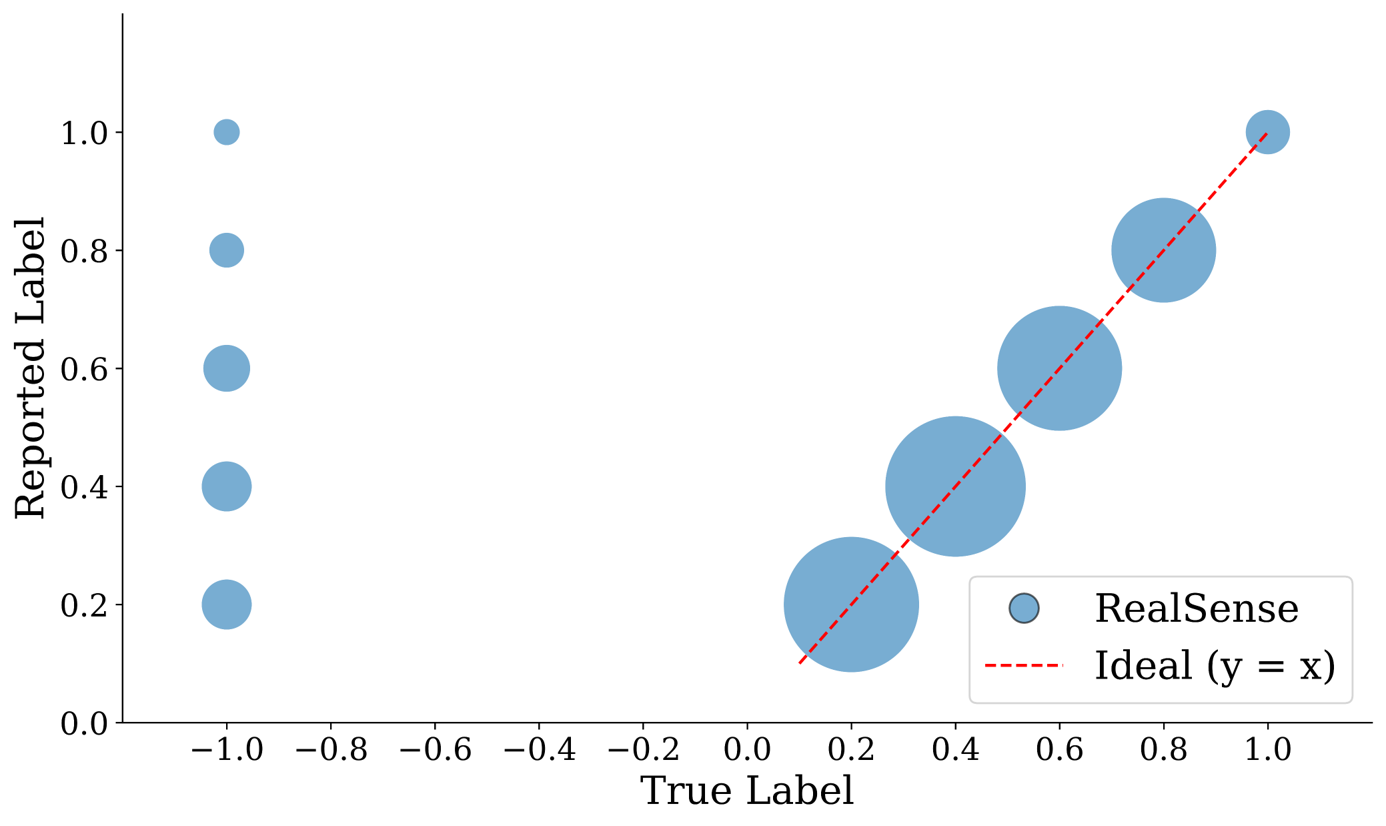}
\caption{Illustration of label discrepancies in GraspNet-1Billion: the horizontal axis shows original GraspNet-1Billion scores, while the vertical axis represents recalculated quality. Many grasps labeled ``valid'' by GraspNet (upper right region) are deemed collision-prone or infeasible upon reevaluation.}
\label{fig:graspnet_mismatch}
\end{figure}

\begin{figure}[t]
\centering
\includegraphics[width=0.9\linewidth]{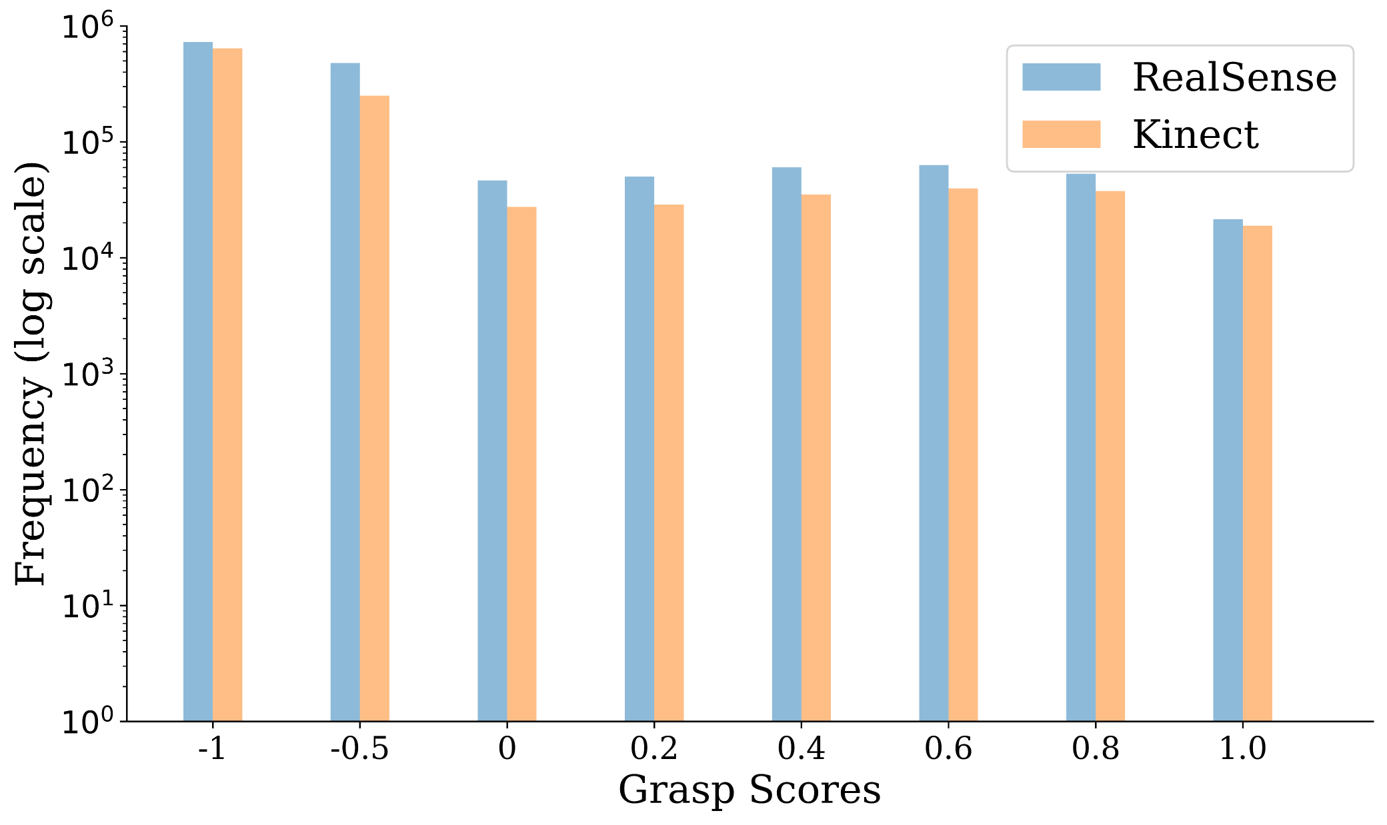}
\caption{Histogram of the curated dataset's grasp score distribution, separated by sensor type (RealSense and Kinect). The y-axis is on a logarithmic scale. Scores on the x-axis categorize the grasps: -1.0 indicates a definite collision, -0.5 marks a low-quality or infeasible grasp, and scores from 0.0 to 1.0 represent valid grasps of increasing quality.}
\label{fig:data_hist}
\end{figure}

\subsection{Dataset Generation}
\label{sub:m4}
Accurate training of a grasp quality prediction model requires a dataset composed of diverse, physically meaningful grasp configurations with reliable labels. To this end, the GraspNet-1Billion benchmark~\cite{fang2020graspnet} was used as the foundation. This dataset includes 100 training scenes and 90 test scenes, each captured from 256 viewpoints using both RealSense and Azure Kinect sensors, offering a rich variety of object configurations and camera perspectives.
Grasp quality in this work is quantified based on the minimum friction coefficient $\mu$ required for a stable, force-closure grasp. This coefficient describes the necessary friction at the contact interface to prevent slippage under ideal conditions. A low required $\mu$ indicates a robust grasp geometry, while a high $\mu$ signifies increased dependence on friction and, consequently, a less stable configuration. This physically grounded metric allows finer-grained evaluation than binary success/failure labels.
Upon evaluation of the original grasp annotations provided by GraspNet-1Billion, a significant mismatch was discovered: many poses labeled as “valid” resulted in gripper-object collisions when the full gripper body geometry was taken into account. Figure~\ref{fig:graspnet_mismatch} illustrates this discrepancy, highlighting the risk of using the original annotations without correction.
To generate a consistent and reliable training dataset, a new pipeline was constructed. For each scene, a fresh set of 7-DoF grasp candidates was generated using the GPG, by following the procedure described in Section~\ref{sub:m1}. These candidates were then evaluated using the official GraspNet-1Billion evaluator, which assigns a quality score based on the aforementioned friction coefficient metric.
The following scoring scheme was used to annotate each grasp:
\begin{itemize}
    \item Grasps resulting in any collision were assigned a fixed score of -1.0.
    \item Grasps that were collision-free but required a high friction coefficient to succeed were assigned a score of -0.5, indicating physical feasibility but practical unreliability.
    \item Grasps that passed collision checks and demonstrated force-closure were assigned a continuous score in the range [0.0, 1.0], where higher values correspond to lower required friction (e.g., $\mu \approx 0.1$) and more reliable contact geometry.
\end{itemize}
\noindent
This refined dataset accurately reflects realistic candidate distributions and provides precise supervision in line with physical execution constraints. The score distribution of the final dataset is visualized in Figure~\ref{fig:data_hist}.

\subsection{Training Strategies and Implementation Details} \label{sub:m5}

Ensemble methods are widely recognized for their ability to reduce variance and enhance generalization~\cite{lakshminarayanan2017simple}. In robotic grasping, such techniques have been successfully applied by combining different architectures~\cite{asif2018ensemblenet} or training the same model on distinct data folds~\cite{xia2023ligbind}. This approach is particularly well suited to GtG~2.0, given the lightweight nature of its network—each model in the ensemble contains only 0.11 million parameters. Even when combined, the full ensemble remains significantly smaller than common alternatives like PointNetGPD (1.6M) and GPD (3.6M), while achieving superior robustness.

The final ensemble consists of five models, all sharing the same architecture but initialized with different random seeds. Each model is trained on a different subset of the data, using 90 out of the 100 available GraspNet-1Billion training scenes (annotation ID 0 only), with the remaining 10 scenes reserved for validation. This stratified folding ensures that each model learns slightly different data distributions. From each fold, the model with the lowest Mean Squared Error (MSE) on its validation set is selected for inclusion. At inference time, the final grasp score is computed by averaging the predictions from all five models, effectively smoothing prediction variance and enhancing the model’s ability to generalize across cluttered, real-world scenes.

\subsubsection{Dynamic Data Sampling}

To mitigate the effects of severe class imbalance in the training data, a dynamic sampling strategy was employed. Specifically, the training set was periodically re-sampled every 10 epochs to maintain a more balanced distribution of grasp types. Each refreshed training set included all available feasible (high-quality) grasps, along with 50,000 randomly selected collision grasps and 50,000 randomly selected low-quality, non-colliding grasps.
This approach ensured that the model remained consistently exposed to positive samples throughout training, while avoiding overfitting to the large and redundant pool of negative examples. By rebalancing the data dynamically, the model was encouraged to focus on learning meaningful geometric cues associated with successful grasps, ultimately improving its discriminative capacity across the full quality spectrum.

\subsubsection{Training Details}
The models were implemented using the \texttt{Pytorch}\cite{10.5555/3454287.3455008}, \texttt{PyTorch Geometric} (\texttt{PyG})\cite{fey2019fast}. Each of the five models in the ensemble was trained for 500 epochs, utilizing an Adam optimizer\cite{kingma2014adam} with a learning rate of $0.01$, and an MSE loss function.

\section{Experiments}
\label{sec:experiments}

This section presents a comprehensive evaluation of the proposed GtG~2.0 framework through both large-scale benchmark testing and real-world robotic experiments. First, quantitative results on the GraspNet-1Billion benchmark are reported, comparing GtG~2.0 against a variety of methods. Next, ablation studies are conducted to analyze the contribution of key architectural components. Finally, the framework’s practical effectiveness is validated through real-world trials on a 4-DoF robotic setup, demonstrating its robustness and adaptability in cluttered environments.

\subsection{GraspNet-1Billion Benchmark Evaluation}
\label{sub:benchmark}

\begin{figure*}[t]
    \centering
    \includegraphics[width=\textwidth]{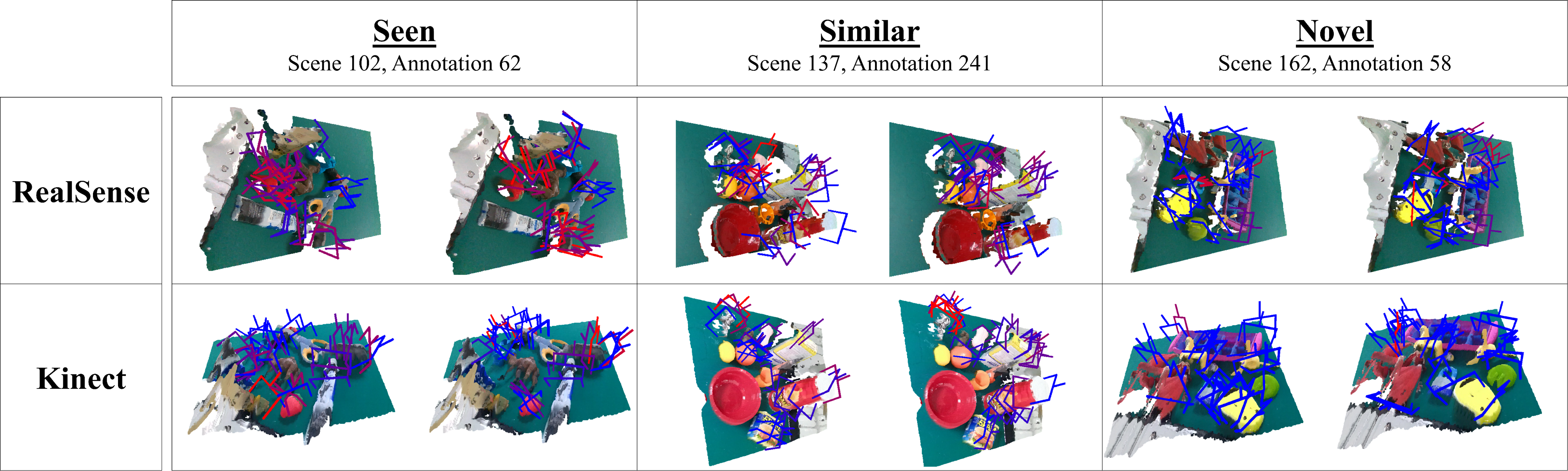}
    \caption{Visualization of the final 50 grasps detected by GtG~2.0 and evaluated by the GraspNet benchmark. Each subimage displays 25 top-scoring grasps overlaid on the scene from different viewpoints. Warmer colors denote higher predicted grasp scores, while cooler colors correspond to lower scores. Splitting the grasps into two groups of 25 improves clarity by preserving object detail compared to plotting all 50 simultaneously.}
    \label{fig:test_eg}
\end{figure*}

The performance of GtG~2.0 was evaluated on the GraspNet-1Billion benchmark, where Average Precision (AP) serves as the primary metric. To ensure full coverage of the scene, a filtering protocol is applied before evaluation. Following the benchmark procedure, grasp pose predictions are first clustered using NMS with thresholds of \(\Delta T = 0.03\,\text{m}\) and \(\Delta \alpha = 30^\circ\). From each cluster, only the grasp with the highest predicted score is retained. To further balance the evaluation, each object is limited to a maximum of 10 grasp proposals. After this filtering stage, the top 50 grasps with the highest predicted scores are selected for evaluation. Each of these grasps is then validated across a range of friction coefficients \(\mu \in \{0.2, 0.4, 0.6, 0.8, 1.0\}\). Precision@k is computed for each \(\mu\), and the final AP is obtained by averaging across all values~\cite{lin2014microsoft}. Figure~\ref{fig:test_eg} illustrates the 50 predicted grasps retained by this process that were ultimately evaluated by the GraspNet benchmark. 

\subsubsection{Comparative Evaluation of Hypothesis-and-Test Methods}

As summarized in Table~\ref{tab:hypothesis_tests_results}, GtG~2.0 consistently outperforms established hypothesis-and-test methods, including GPD and PointNetGPD, on the GraspNet-1Billion benchmark. Across all test splits, GtG~2.0 achieves higher AP while maintaining superior computational efficiency. This efficiency is reflected in the compact size of the model ($0.57$M parameters, compared to $3.6$M for GPD and $1.6$M for PointNetGPD) and in its lean data requirements. Whereas PointNetGPD processes a fixed set of $1000$ points for every candidate grasp, GtG~2.0 relies on a flexible representation of at most $140$ points, capturing both \emph{inside} and \emph{outside} regions. By combining improved accuracy with greater efficiency, GtG~2.0 establishes itself as a state-of-the-art hypothesis-and-test framework for grasp pose detection.

\begin{table}[t]
\centering
\setlength{\tabcolsep}{6pt} %
\renewcommand{\arraystretch}{1.15} %
\caption{Results on GraspNet-1Billion Dataset for Hypothesis-and-Test methods, showing Average APs of each Scene Type on RealSense/Kinect Split.}
\label{tab:hypothesis_tests_results}
\begin{tabular}{lccc}
\toprule
\textbf{Method} & \textbf{Backbone} & \textbf{\#Params} & \textbf{Average (APs(\%))\(\uparrow\)} \\
\midrule
GPD~\cite{gualtieri_high_2016} & CNN      & 3.6M  & 17.48 / 19.05 \\
PointNetGPD~\cite{liang_pointnetgpd_2019} & PointNet  & 1.6M  & 19.29 / 20.88 \\
\textbf{GtG~2.0 (Ours)}          & GNN      & \textbf{0.57M} & \textbf{53.42 / 47.00} \\
\bottomrule
\end{tabular}
\end{table}

\subsubsection{Comparative Evaluation with Other GNN-based Methods}
\label{subsub:gnn_comparison}

Table~\ref{tab:gnn_res} reports a comparison between GtG~2.0 and recent GNN-based approaches, including GraNet and the parallel graph network proposed in~\cite{ZHUANG2025102879}. Across all evaluation metrics, GtG~2.0 achieves substantially higher performance. This improvement underscores the benefit of incorporating both \emph{inside} and \emph{outside} points to capture the local grasp region along with its immediate context. In contrast, prior GNN-based methods often emphasize scene-level reasoning, which provides broader but less discriminative cues for individual grasps. By balancing detailed local geometry with contextual information, GtG~2.0 produces more accurate grasp quality predictions and establishes a stronger baseline for graph-based grasp detection.

\begin{table}[t]
\centering
\setlength{\tabcolsep}{6pt} %
\renewcommand{\arraystretch}{1.15} %
\caption{Comparison of the GNN-based Grasp Pose Detection Methods on the GraspNet-1Billion Benchmark, Showing Average APs of each Scene Type on RealSense/Kinect Split.}
\label{tab:gnn_res}
\begin{tabular}{lcc}
\toprule
\textbf{Method} & \textbf{Category} & \textbf{Average (APs(\%))\(\uparrow\)} \\
\midrule
GraNet~\cite{GraNet} & End-to-End & 32.73 / 29.44 \\
Zhuang et al.~\cite{ZHUANG2025102879} & End-to-End & 36.99 / 32.88 \\
\textbf{GtG~2.0 (Ours)} & Hypothesis-and-Test & \textbf{53.42 / 47.00} \\
\bottomrule
\end{tabular}
\end{table}

\subsubsection{Comparative Evaluation of All Grasp Detection Methods}
\label{subsub:all_methods}

As summarized in Table~\ref{tab:graspnet_results}, GtG~2.0 ranks among the top three methods on the GraspNet-1Billion benchmark, even when compared against recent end-to-end architectures. This result demonstrates the strong potential of GNNs for grasp pose detection in cluttered scenes. The high performance of GtG~2.0 stems from its ability to combine the strengths of both paradigms: the flexibility of hypothesis-and-test pipelines and the enhanced representational power of GNNs, which leverage both \emph{inside} and \emph{outside} points to provide a richer grasp representation.

\begin{table}[t]
  \centering
  \caption[GraspNet-1Billion Results]{%
    Results of All Methods on the GraspNet-1Billion (APs on RealSense/Kinect split).\\ 
    Color-code ranking: 
    \textcolor{softred}{\textbf{Red}}=1st, 
    \textcolor{softgreen}{\textbf{Green}}=2nd, 
    \textcolor{softblue}{\textbf{Blue}}=3rd.}
  \label{tab:graspnet_results}
\resizebox{\linewidth}{!}{%
  \begin{tabular}{lccccc}
    \toprule
    \textbf{Method} & \textbf{Seen\(\uparrow\)} & \textbf{Similar\(\uparrow\)} & \textbf{Novel\(\uparrow\)} & \textbf{Average\(\uparrow\)} & \textbf{Params\(\downarrow\)} \\
    \midrule
    GPD~\cite{gualtieri_high_2016} 
      & 22.87 / 24.38 & 21.33 / 23.18 &  8.24 /  9.58 & 17.48 / 19.05 & 3.6M \\
    PointnetGPD~\cite{liang_pointnetgpd_2019} 
      & 25.96 / 27.59 & 22.68 / 24.38 &  9.23 / 10.66 & 19.29 / 20.88 & 1.6M \\
    RGB Matters~\cite{gou_rgb_2021} 
      & 27.98 / 32.08 & 27.23 / 30.40 & 12.25 / 13.08 & 22.49 / 25.19 & -- \\
    REGNet~\cite{zhao_regnet_2021} 
      & 37.00 / 37.76 & 27.73 / 28.69 & 10.35 / 10.86 & 25.03 / 25.77 & -- \\
    TransGrasp~\cite{10.1109/ICRA46639.2022.9812001} 
      & 39.81 / 35.97 & 29.32 / 29.71 & 13.83 / 11.41 & 27.65 / 25.70 & -- \\
    GraNet~\cite{GraNet} 
      & 43.33 / 41.48 & 39.98 / 35.29 & 14.90 / 11.57 & 32.73 / 29.44 & -- \\
    Scale Balanced Grasp~\cite{ma_towards_2022} 
      & 63.83 / -- & 58.46 / -- & 24.63 / -- & 48.97 / -- & -- \\
    HGGD~\cite{chen_efficient_2023} 
      & 59.36 / 60.26 & 51.20 / 48.59 & 22.17 / 18.43 & 44.24 / 42.43 & 3.42M \\
    GSNet~\cite{wang_graspness_2024} 
      & 67.12 / \third{63.50} & 54.81 / 49.18 & 24.31 / 19.78 & 48.75 / 44.15 & 15.4M \\
    RNGNet~\cite{chen2024regionaware} 
      & \first{76.28} / \first{72.89} & \first{68.26} / \first{59.42} & \first{32.84} / \first{26.12} & \first{59.13} / \first{52.81} & 3.66M \\
    \textit{Zhuang, C. et al}.~\cite{ZHUANG2025102879} 
      & 50.12 / 45.30 & 43.90 / 40.00 & 16.94 / 13.33 & 36.99 / 32.88 & -- \\
    FlexLoG~\cite{XIE2026112088} 
      & \second{72.81} / \second{69.44} & \second{65.21} / \second{59.01} & \second{30.04} / \third{23.67} & \second{56.02} / \second{50.67} & -- \\
    \midrule
    \textbf{GtG~2.0 (Ours)} 
      & \third{68.79} / 62.61 & \third{61.71} / \third{53.93} & \third{29.75} / \second{24.45} & \third{53.42} / \third{47.00} & \textbf{0.57M} \\
    \bottomrule
  \end{tabular}%
  }
\end{table}

\subsubsection{Ablation Studies}
\label{subsub:ablation}

A series of ablation studies were conducted to evaluate the contribution of key components within GtG~2.0. For this analysis, a fixed subset of annotation IDs (0–255 in steps of 10) was selected from the RealSense split. Table~\ref{tab:ablation_studies} reports results for three configurations: (i) using only \emph{inside points}, similar to GPD and PointNetGPD; (ii) incorporating both \emph{inside} and \emph{outside points} to provide contextual information around the grasp region; and (iii) applying the ensemble strategy described in Section~\ref{sub:m5}.  
The inclusion of outside points improved the average AP from 45.29\% to 48.57\%, demonstrating the value of richer contextual cues in accurately identifying graspable regions. Performance was further elevated to 53.69\% with the ensemble strategy, highlighting its effectiveness in mitigating individual model errors through prediction averaging. These results confirm that both contextual point inclusion and model ensembling contribute substantially to the robustness and accuracy of GtG~2.0.

\begin{table}[t]
\centering
\caption{Ablation Studies on Different Components. Results obtained from a fixed subset of annotation IDs ranging from 0 to 255 in increments of 10, for the RealSense Split.}
\label{tab:ablation_studies}
\renewcommand{\arraystretch}{1.3} %
\begin{tabular}{lc}
\toprule
\textbf{Configuration} & \textbf{Average (APs(\%))\(\uparrow\)} \\
\midrule
Inside Points Only & 45.29 \\
Inside \& Outside Points & 48.57 \\
Ensemble of 5 Models & 53.69 \\
\bottomrule
\end{tabular}
\end{table}

\subsection{Physical Robot Experiments}
\label{sec:realworld_pipeline}

\begin{figure*}[t]
\centering
\includegraphics[width=\textwidth]{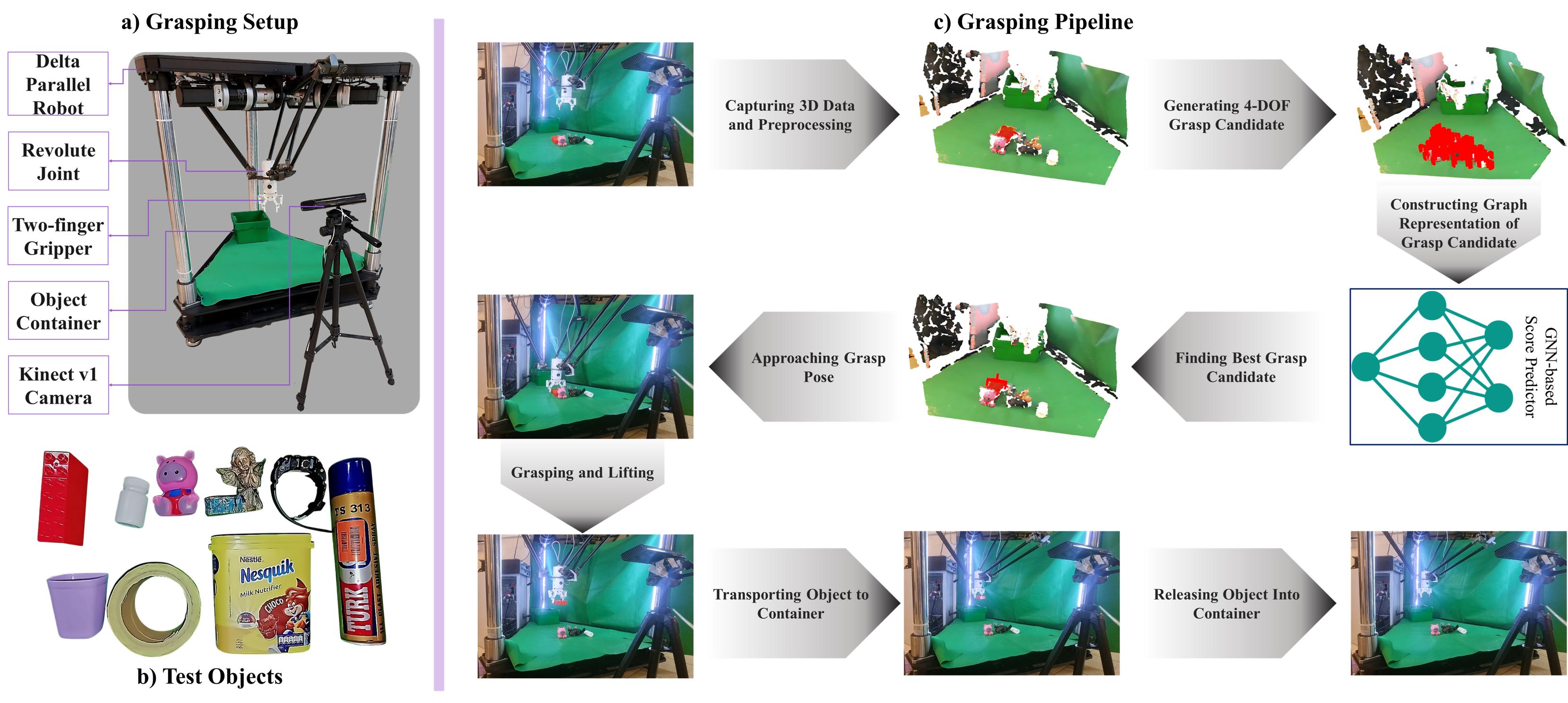}

    \caption{Overview of the physical robot experiments. 
\textbf{(a)} Experimental setup with an overall 4-DoF configuration, consisting of a 3-DoF Delta Parallel robot equipped with a two-finger gripper featuring an additional rotational DoF around the $z$-axis, along with a Kinect~v1 sensor. 
\textbf{(b)} Set of nine household objects with diverse shapes, sizes, and materials used to create cluttered test scenarios. 
\textbf{(c)} Execution of the grasping pipeline, including point cloud acquisition, heuristic grasp candidate generation, GNN-based scoring, and final object placement into a container.}

\label{fig:allset}
\end{figure*}

While the GraspNet-1Billion benchmark provides a comprehensive simulation-based evaluation, real-world validation is essential to demonstrate practical applicability. For this purpose, GtG~2.0 was tested on a robotic platform with an overall 4-DoF grasping capability. The setup consists of a 3-DoF Delta Parallel robot combined with a two-finger gripper featuring a rotational degree of freedom around its $z$-axis, yielding a total of 4 DoFs for top-down grasping~\cite{yarmohammadi2023experimental}. The system was developed at the Human and Robot Interaction Laboratory, University of Tehran, and was inspired by the design of the Robotiq gripper~\cite{demers2015gripper}. The experimental setup is illustrated in Figure~\ref{fig:allset}(a). A Kinect~v1 camera was employed for point cloud acquisition, and a custom 4-DoF grasp candidate generator was implemented, since the 7-DoF GPG is not directly compatible with this 4-DoF configuration. The full testing procedure is outlined below.

\begin{enumerate}
    \item \textbf{Test Objects and Scene Setup:}  
    Nine household objects with diverse shapes, sizes, and materials were selected for evaluation (Figure~\ref{fig:allset}(b)). In each trial, a random subset of 5--8 objects was placed within the robot’s workspace to create cluttered scenes.

    \item \textbf{Point Cloud Acquisition and Processing:}  
    A Kinect~v1 camera was used to capture raw point clouds of the workspace. The data were transformed into the robot global frame, denoised using a weighted \(k\)-nearest neighbors filter \((k=10)\), and downsampled with a voxel size of 2\,mm.

    \item \textbf{Heuristic 4-DoF Candidate Generation:}  
    Since the original 7-DoF generator is not directly compatible with the 4-DoF setup, a heuristic generator was developed. The grasp space was discretized by subdividing \((x, y, z)\) coordinates into 1\,cm intervals. To mitigate gaps caused by occlusions, each point was replicated at successively lower \(z\)-values down to a minimum height. Candidates' orientations were discretized in 30° increments around the \(z\)-axis, and a fixed 10\,cm gripper width was assigned to all grasps.

    \item \textbf{Grasp Scoring and Execution:}  
    For each candidate, a graph representation was constructed and evaluated by the trained GNN ensembles. The grasp pose with the highest predicted score was selected and executed by the robot.
\end{enumerate}

Following this pipeline, illustrated in Figure~\ref{fig:allset}(c), GtG~2.0 achieved an overall success rate of 91\% across five test scenes, each containing 5--8 objects. All scenes were completely cleared, resulting in a 100\% completion rate. The outcomes are summarized in Table~\ref{table:robotics_experiments}, demonstrating the robustness and adaptability of GtG~2.0 under different sensor configurations and with a generator constrained to 4-DoF.

\begin{table}[t]
    \caption{Results of the 4-Dof robot grasping experiments conducted using a 3-Dof 
    Delta Parallel robot equipped with a two-finger gripper and a Kinect-v1 sensor in cluttered scenarios, reporting for each scene the number of objects present, the number of successful grasps achieved, and the total number of grasp attempts.}
    \label{table:robotics_experiments}
    \centering
    \begin{tabular}{c c c c}
    \hline
    \textbf{Scene} & \textbf{\#Objects} & \textbf{Success} & \textbf{Attempts} \\
    \hline
    1 & 5 & 5 & 5 \\
    2 & 5 & 5 & 6 \\
    3 & 6 & 6 & 7 \\
    4 & 8 & 8 & 8 \\
    5 & 6 & 6 & 7 \\
    \hline
    \\[-0.8em]
    \multicolumn{2}{l}{\textbf{Success Rate}} & \multicolumn{2}{r}{30 / 33 = \textbf{91\%}} \\[1em]
    \multicolumn{2}{l}{\textbf{Completion Rate}} & \multicolumn{2}{r}{5 / 5 = \textbf{100\%}} \\
    \\[-0.8em] 
    \hline
    \end{tabular}
\end{table}

\section{Conclusion} 
\label{sec:conclusion}

This paper introduced \emph{Grasp the Graph 2.0 (GtG~2.0)}, a lightweight hypothesis-and-test framework for high-precision grasp pose detection in cluttered environments. The framework leverages an ensemble of Graph Neural Networks with a novel representation that incorporates both \emph{inside points} and \emph{outside points}, enabling richer geometric reasoning than conventional methods. Experiments on the GraspNet-1Billion benchmark demonstrated that GtG~2.0 achieves state-of-the-art performance among hypothesis-and-test and GNN-based approaches, while ranking among the top three methods overall. The framework achieves up to 35\% higher Average Precision than comparable baselines, despite requiring up to six times fewer parameters.  
Physical experiments on a 4-DoF Delta robot with a Kinect-v1 sensor further validated the approach, yielding a 91\% grasp success rate and 100\% scene completion, underscoring its robustness to new sensors and candidate generators. Nevertheless, limitations remain: performance on novel scenes lags behind seen scenarios, and the grasp candidate generator, while efficient, still produces a large fraction of low-quality grasps. Addressing these challenges points to opportunities for future work, including the design of more advanced candidate generation strategies, improving generalization to unseen environments, and extending GtG~2.0 into a single-stage end-to-end framework.  
Overall, the results establish GtG~2.0 as a practical and effective approach to robotic grasp detection, combining the flexibility of hypothesis-and-test pipelines with the efficiency and expressiveness of GNN-based reasoning.

\section{Declaration of competing interests}
The authors declare that they have no known competing financial interests or personal relationships that could have appeared to influence the work reported in this paper.

\section{Declaration of generative AI and AI-assisted technologies in the writing process}

During the preparation of this work, the authors used ChatGPT and Grammarly in order to improve language and readability. After using these tools/services, the authors reviewed and edited the content as needed and take full responsibility for the content of the publication.

\bibliographystyle{elsarticle-num}
\bibliography{main}

\clearpage

\end{document}